\documentclass[runningheads]{llncs}
\usepackage[T1]{fontenc}
\usepackage{graphicx,xcolor}
 \usepackage{algpseudocode, algorithm,amsmath}
\usepackage{amssymb}
\usepackage{amsfonts}

\begin{document}
\title{Expected Free Energy-based Informative Path Planning for Robotic Mars Exploration }
\titlerunning{EFE based Informative Path Planning}
%

\author{Ajith Anil Meera\inst{1}, Pablo Lanillos \inst{2} and Wouter Kouw \inst{1}}
\authorrunning{A. Anil Meera et al.}
%
\institute{ TU Eindhoven, The Netherlands \and Cajal Centre for Neuroscience, Spanish National Research Council, Madrid, Spain}
%
\maketitle              
\begin{abstract}
An autonomous robot efficiently exploring an unknown environment, such as looking for water sources on Mars, faces two simultaneous demands: building an accurate information map while quickly finding the regions of greatest value, and paying for every meter of travel and the cost of every measurement it takes. Classical information-seeking and reward-seeking criteria address only one of these objectives at a time. Here, we propose Expected Free Energy (EFE), the principled action-selection objective from active inference, as a unifying criterion for budgeted robotic informative path planning. Maintaining a Gaussian-process belief over the information field, our agent plans continuous trajectories that minimize expected free energy under hard path-length constraints. The results from multiple realizations show that EFE-based planning yields accurate posterior maps and locates the highest-value regions simultaneously, outperforming information-theoretic baselines under the same settings. In robotic exploration, these unified, easy-to-tune principled information-gathering strategies facilitate autonomous deployment while enforcing efficiency and resource constraints.


\end{abstract}
\section{Introduction}
A robot deployed to explore and find water sources, monitor an oil spill, map a radiation field, survey a crop \cite{popovic2020informative}, locate cyanobacteria \cite{hitz2017adaptive}, or find survivors~\cite{meera2019obstacle} must answer two questions at once: where is the phenomenon strongest and what does the information field look like? Every measurement costs travel,
energy, and time, and in real-world settings, the robot should adhere to its resource limitations. Thus, the robot can only visit and perform measurements on a few of the infinitely many
locations available in the environment. Informative Path Planning (IPP) addresses this by selecting sampling trajectories that maximize the value of the data collected under a movement budget~\cite{hollinger2014sampling,chen2024adaptive}. The central difficulty
is the exploration--exploitation trade-off: the robot must spend its limited
budget both on reducing uncertainty about the field and on pinpointing its most
valuable regions.

Existing criteria address only one side of this trade-off. Information-theoretic
objectives such as mutual information \cite{bai2016information} or variance reduction \cite{popovic2020informative} drive pure
exploration and yield accurate maps, but ignore where value lies. Bayesian-optimization acquisitions, such as expected improvement and upper confidence bound~\cite{srinivas2012information}, drive exploitation toward the optimum, but neglect the rest of the map~\cite{shahriari2015taking,williams2006gaussian}. Practitioners typically pick one or hand-tune a weighted sum, with no principled way to balance the two.

\begin{figure}
\includegraphics[width=0.94\textwidth]{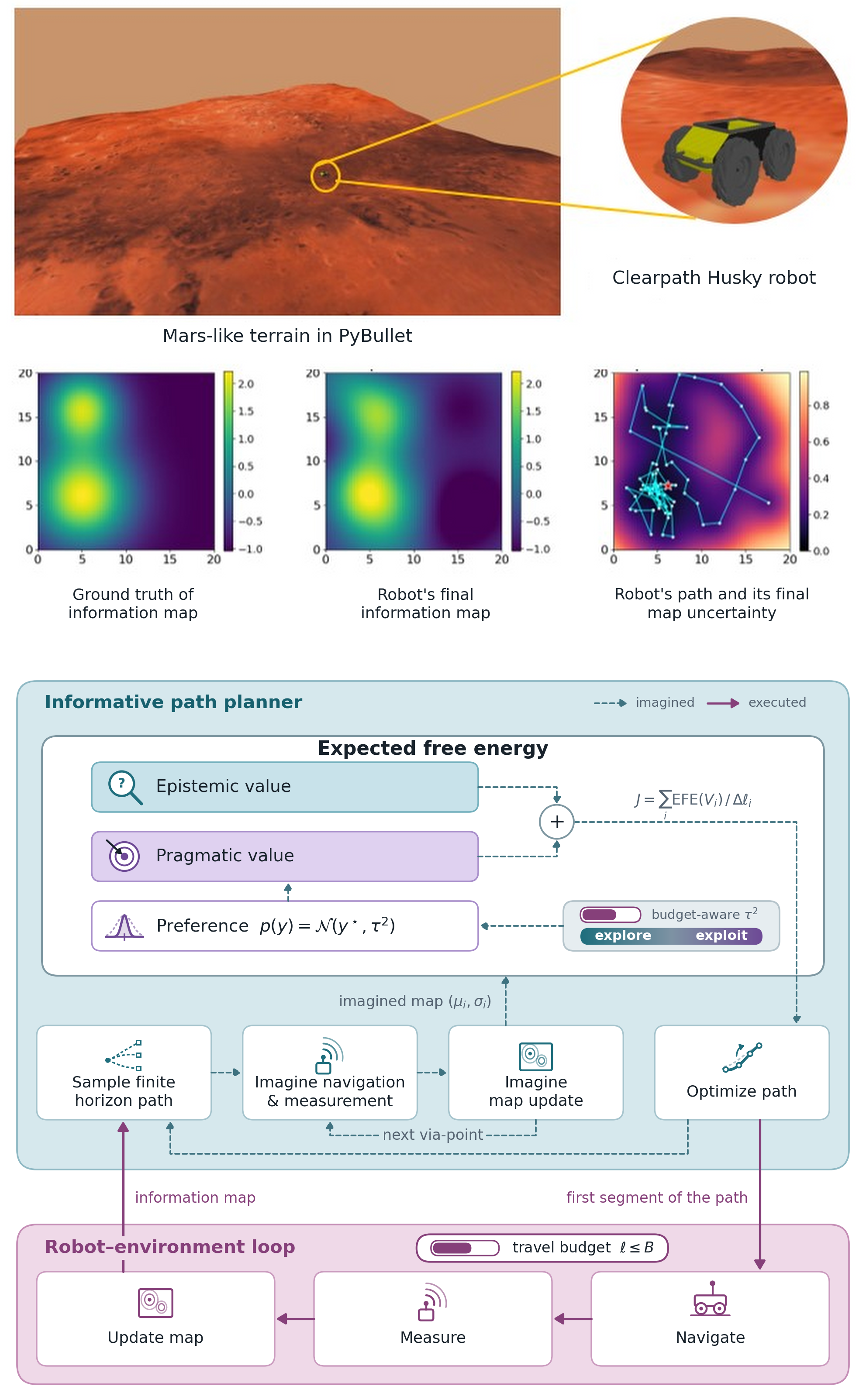}
\caption{\textbf{Proposed EFE-IPP algorithm and simulation setup and one exploration result}. The top panel shows the Mars-like simulation environment used in PyBullet with a Clearpath Husky robot. The middle plot shows the ground truth information map, the robot's final information map, its uncertainty and the robot's path. The bottom panel shows our EFE-IPP algorithm to optimize the path.} \label{teaser_ipp}
\end{figure}

Active inference \cite{friston2010free} offers a principled alternative. Originating as a theory of how
the brain perceives and acts, it casts behavior as the minimization of expected
free energy (EFE), a single objective that decomposes into a goal-seeking
(pragmatic) term and an information-seeking (epistemic) term~\cite{da2020active,parr2022active}.
EFE thus resolves exploration versus exploitation not by an arbitrary weighting
but as a consequence of one coherent, brain-inspired criterion. This makes EFE a
natural candidate for robotic information gathering~\cite{wakayama2023active}. 
Despite the potential application of active inference in robotic systems~\cite{lanillos2021active,da2022active}, its use in navigation is still under development. For example, it has been centered on discrete and semantic navigation~\cite{kaplan2018planning,ccatal2021robot,taniguchi2023active,de2024exploring,de2025navigation}, and autonomous driving under imitation learning schemes~\cite{nozari2022active}. More recently, EFE has been used within Bayesian optimization to myopically select the next best query point~\cite{meera2026curvature}. How EFE should drive a mobile robot that plans continuous trajectories subject to a hard travel budget remains an open problem. Existing active-inference planners largely operate either on discrete
Markov decision processes \cite{friston2021sophisticated} or on high-dimensional control with learned policies \cite{fountas2020deep,tschantz2020reinforcement}, leaving the budgeted, continuous-trajectory IPP setting underexplored. We fill this
gap by introducing an IPP algorithm that i) plans continuous trajectories online, ii) requires no policy enumeration or offline training, and iii) respects a hard travel budget. Figure \ref{teaser_ipp} shows the architecture of our proposed IPP algorithm. The core contributions of this work are:
\begin{itemize}
\item We introduce an adaptive, non-myopic, computationally tractable,  EFE-based IPP framework that plans continuous trajectories in a receding-horizon loop, under  a hard travel budget.
\item We introduce a budget-aware annealing schedule for EFE that adaptively
      balances exploration and exploitation using the fraction of budget consumed.
  \item We show that the proposed planner outperforms information-theoretic
        and Bayesian-optimization baselines on map accuracy and simple
        regret.
\end{itemize}

\section{Problem Statement}
A robot explores a bounded region $\Omega \subset \mathbb{R}^2$ in search of
an unknown scalar information field $f : \Omega \to \mathbb{R}$ that it cannot
observe directly. The only way the robot learns $f$ is by traveling to a
location and taking a noisy measurement there. Following a continuous path
$\gamma$, the robot collects $N$ samples at chosen positions
$X_1, \dots, X_N \in \Omega$, each corrupted by additive sensor noise,
\begin{equation} \label{eqn:observation}
  y_i = f(X_i) + \varepsilon_i, \qquad
  \varepsilon_i \sim \mathcal{N}(0, \sigma_n^2),
\end{equation}
with noise variance $\sigma_n^2 > 0$. Travel is not free: a movement budget
$B > 0$ caps the path's total length $\ell(\gamma)$. The challenge is to
spend this budget well, choosing where to travel and where to sample so that
the collected measurements are as informative as possible. In this work,
``informative'' means two things at once: the measurements should let the
robot reconstruct an accurate map of $f$ across $\Omega$, \emph{and} locate
the field's most valuable region. Letting $J$ be a utility that scores both,
the informative path planning (IPP) problem is
\begin{equation}
  \max_{\gamma,\; X_{1:N}}\;
  J\!\left(\{(X_i, y_i)\}_{i=1}^{N}\right)
  \quad \text{s.t.} \quad
  \ell(\gamma) \le B, \;\; \gamma \subset \Omega .
\end{equation}

\section{Method}
We maintain a probabilistic belief over the field, plan short trajectories
that maximize expected free energy per unit travel, and execute them in a
receding-horizon loop until the movement budget is spent. Algorithm~\ref{alg:efe-ipp}
summarizes the procedure.

\begin{algorithm}[!htb]
\caption{EFE-driven informative path planning}
\label{alg:efe-ipp}
\begin{algorithmic}[1]
\Require movement budget $B$, planning horizon $n$
\Ensure GP belief $\mathcal{G}$ over the field
\State $\mathcal{G} \gets$ \Call{InitGP}{}
       \Comment{fit GP to a few initial samples}
\State $x \gets$ start pose,\quad $\ell \gets 0$
\While{$\ell < B$}
    \State $\gamma^{\star} \gets \arg\max_{\gamma}$
           \Call{EvaluatePath}{$\gamma, \mathcal{G}, x$}
           \Comment{plan the path: DE, $\gamma$ feasible}
    \State $v \gets$ \Call{MoveRobot}{$\gamma^{\star}$}
           \Comment{execute one step of the path}
    \State $y \gets$ \Call{TakeMeasurement}{$v$}
           \Comment{noisy reading at $v$}
    \State $\mathcal{G} \gets$ \Call{UpdateGP}{$\mathcal{G}, v, y$}
           \Comment{fold the reading into the belief}
    \State $x \gets v$,\quad $\ell \gets \ell + \text{step length}$
\EndWhile
\State \Return $\mathcal{G}$
\Statex
\Function{EvaluatePath}{$\gamma, \mathcal{G}, x$}
    \State $\mathcal{G}_f \gets \mathcal{G}$,\quad $S \gets 0$
           \Comment{$\mathcal{G}_f$: a hypothetical look-ahead belief}
    \For{each waypoint $v_i$ along $\gamma$}
        \State $(\mu_i, \sigma_i) \gets$ \Call{Predict}{$\mathcal{G}_f, v_i$}
               \Comment{GP posterior at $v_i$}
        \State $S \gets S + \mathrm{EFE}(\mu_i, \sigma_i)\,/\,\text{step length}_i$
               \Comment{reward per metre}
        \State $\mathcal{G}_f \gets$ \Call{UpdateGP}{$\mathcal{G}_f, v_i, \mu_i$}
               \Comment{fantasy: treat prediction as the reading}
    \EndFor
    \State \Return $S$
\EndFunction
\end{algorithmic}
\end{algorithm}
The following sections will explain the methodology followed by each subprocess within the algorithm.

\subsection{Modeling the Information Map with a Gaussian Process}
We model the unknown field $f$ as a GP. A GP fits this setting
for three reasons: (i) it admits a closed-form posterior from a handful of
noisy point measurements, (ii) it returns predictive uncertainty at every location, and (iii) it stays sample-efficient under the small data regimes
typical of budget-constrained exploration. After $t$ measurements have been collected into the dataset
$\mathcal{D}_t = \{(X_i, y_i)\}_{i=1}^{t}$, the GP posterior is Gaussian at every location, with predictive mean $\mu_t(X)$ and variance $\sigma_t^2(X)$ available in closed form. We adopt the posterior mean as the robot's field estimate, $\hat{f} \equiv \mu_t$. The robot never observes the full $f$ directly. It must reconstruct it online from its own samples, and this GP belief is the only summary of the environment the planner uses.

\subsection{Expected Free Energy as the Objective Function for Planning}
Each candidate measurement position $X$ is scored by its (negative) expected free energy
(EFE), which decomposes into a pragmatic value and an epistemic value \cite{friston2015active}:
\begin{equation}
  \mathrm{EFE}(X) =
  -\underbrace{\mathbb{E}_{q(y\mid X)}\!\left[-\ln p(y)\right]}_{\text{pragmatic value}}
  +\underbrace{\mathbb{E}_{q(y\mid X)}\!\left[D_{\mathrm{KL}}\!\big(q(f\mid y,X)\,\|\,q(f\mid X)\big)\right]}_{\text{epistemic value}} .
  \label{eq:efe-generic}
\end{equation}
The pragmatic term is the expected log-loss of the predicted reading $y$ against the agent's preference distribution $p(y)$ and the epistemic term is the expected reduction in posterior uncertainty about $f$ induced by the measurement. Under the GP belief and a Gaussian preference
$p(y) = \mathcal{N}(y^{\star}, \tau^2)$, both terms admit a closed form
\cite{meera2026curvature},
\begin{equation}
  \mathrm{EFE}(X) =
  -\underbrace{\frac{\big(\mu_t(X) - y^{\star}\big)^{2}}{2\tau^{2}}
   -\,\frac{\sigma_t^{2}(X)}{2\tau^{2}}}_{\text{pragmatic value}}
  +\underbrace{\,\frac{1}{2}\log\!\Big(1 + \frac{\sigma_t^{2}(X)}{\sigma_n^{2}}\Big)}_{\text{epistemic value}} .
  \label{eq:efe}
\end{equation}
Here $\sigma_n^{2}$ is the observation noise variance from
Equation~\ref{eqn:observation}, and $\tau^{2} > 0$ encodes the variance of the preference $p(y)$. Setting $y^*$ to be high encourages seeking high observations. However, since $y^*$ is unknown, we select the current best observation with an optimistic offset through $y^{\star} = y_{\text{best}} + \kappa$.  Here, $y_{\text{best}} = \max_{i \le t} y_i$ is the best reading so far, $\kappa > 0$ is an aspiration offset that sets what the agent wants to see. The pragmatic value rewards positions whose predicted reading is close to $y^{\star}$ and whose prediction the agent already trusts (driving exploitation). The epistemic value rewards positions whose observations would most reduce posterior uncertainty about $f$ (driving exploration). The temperature $\tau^{2}$ alone arbitrates the two: a single knob that lets the same objective span pure exploration, pure optimization, or any trade-off between them. Therefore, maximizing EFE encourages high observations (exploitation) and balances it with the information gain (exploration) through a single temperature parameter $\tau^2$.

\subsection{Budget-Aware Balance of Exploration and Exploitation}
The right balance between exploration and exploitation shifts over a
mission: early on the map is poor and information is worth more, while late on the remaining budget is too small to exploit fresh information and is better spent securing value. We track this progress as the fraction of budget consumed,
\begin{equation}
  p \;=\; \ell / B \;\in\; [0, 1],
\end{equation}
where $\ell$ is the distance traveled so far. We introduce a budget-aware update  rule for $\tau^2$. As $p$ grows, we cool the temperature $\tau^{2}$ smoothly from a high value $\tau^{2}_{\max}$ (favouring exploration) toward a low value $\tau^{2}_{\min}$ (favouring exploitation),
\begin{equation} 
  \tau^{2}(p) \;=\; \tau^{2}_{\min}
  + \frac{\tau^{2}_{\max} - \tau^{2}_{\min}}{1 + e^{\,k\,(p - p_0)}},
  \label{eq:tau-schedule}
\end{equation}
where $p_0 \in (0,1)$ is the budget fraction at which $\tau^{2}$ has dropped halfway between the two extremes, and $k > 0$ controls how sharply the transition happens around that point. As $p$ crosses $p_0$, the exponent flips sign and the schedule swings from $\tau^{2} \approx \tau^{2}_{\max}$ to $\tau^{2} \approx \tau^{2}_{\min}$. The agent opens its mission with exploration and closes it with exploitation, within the same objective and the same budget. Although Equation \ref{eq:tau-schedule} introduces multiple constants, $\tau_{max}^2$ and $\tau_{min}^2$ are the main ones and can easily be tuned to represent sufficiently high exploration and exploitation, respectively.

\subsection{Trajectory Objective via a Fantasy Path}
The robot moves along a continuous trajectory. So we parameterize a candidate plan as a smooth polynomial path through its current position $X$ and $n$ via-points $V_1, \dots, V_n \in \Omega$, where $n$ is the planning horizon. We score the whole plan by its total EFE per metre travelled,
\begin{equation}
  J(V_{1:n}) \;=\; \sum_{i=1}^{n}
  \frac{\mathrm{EFE}(V_i)}{\max\!\big(\Delta\ell_i,\,\delta\big)},
  \label{eq:traj-obj}
\end{equation}
where $\Delta\ell_i$ is the arc length of the $i$-th segment and $\delta > 0$ floors vanishingly short segments so that they cannot dominate the objective. Evaluating $\mathrm{EFE}(V_i)$ requires a belief about $f$ at $V_i$. But during planning, no measurements have actually been taken. We supply this belief through a \emph{fantasy path}: at each via-point, we imagine the measurement to equal the GP's predicted mean $\mu(V_i)$, update the GP with that imagined sample, and use the updated belief to score the next via-point. The path score $J$ is then the sum of these EFE values along the fantasy path, so the planner correctly anticipates that uncertainty shrinks as the robot advances.

\subsection{Receding-Horizon Execution}
We execute the plan in a receding-horizon, model-predictive-control (MPC) loop: the planner looks $n$ steps ahead at every cycle but commits only to the first segment, replanning against the updated belief after each measurement. This makes the planner non-myopic, since it scores entire trajectories rather than isolated points, and adaptive, since every new measurement feeds the next plan. At each cycle the planner finds the optimal via-points:
\begin{equation}
  V^{\star}_{1:n} \;=\;
  \arg\max_{V_{1:n} \in \Omega^{n}} \; J(V_{1:n})
  \quad \text{s.t.} \quad
  \Delta\ell_1 \le L_{\max}, \;\;
  \sum_{i=1}^{n} \Delta\ell_i \le B - \ell,
  \label{eq:plan}
\end{equation}
where $L_{\max} > 0$ is a per-segment kinematic cap that prevents the robot from teleporting between samples, and $B - \ell$ is the remaining budget. The robot then drives along the first spline segment from $X$ to $V^{\star}_1$, takes a noisy measurement $y$ there, and conditions the GP on $(V^{\star}_1, y)$. The GP hyperparameters are refit periodically. The cycle repeats from the new pose with the updated belief until $\ell$ reaches $B$.

\begin{figure}[!t]
\includegraphics[width=\textwidth]{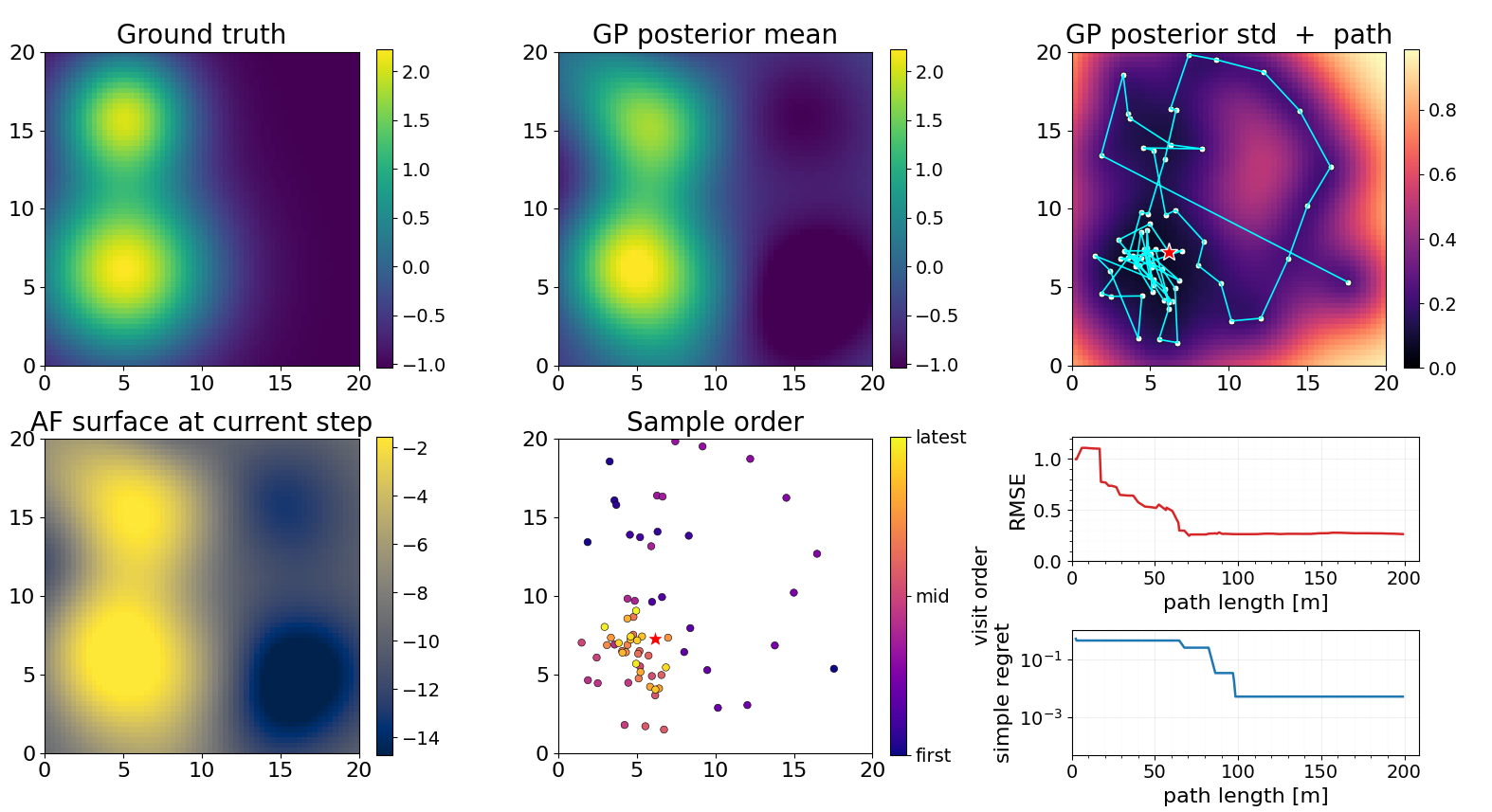}
\caption{The top panel shows the ground truth of the information map, the final GP posterior mean and uncertainty, after following the path shown in the top right panel. The bottom panel shows the EFE objective function, the order of next locations selected for information gathering, the RMSE of the information map and the simple regret plotted against robot path length.} \label{ipp_result}
\end{figure}

\subsection{Path Optimization through Evolutionary Algorithm}
At each cycle the planner picks the $n$ via-points $V_{1:n}$ that maximize the per-metre EFE objective in Eq.~\eqref{eq:plan}. The fantasy-path scoring approximates the otherwise intractable belief-space POMDP look-ahead with a single deterministic objective. This reduces planning to a non-convex, $2n$-dimensional continuous optimization problem in which $J$ is cheap to evaluate, but might have no useful gradient in practice. Therefore, we solve it with Differential Evolution (DE) \cite{das2010differential}, a population-based evolutionary optimizer suited to gradient-free, non-convex continuous problems. At every generation, DE evaluates a population of candidate paths under $J$, then mutates and recombines the higher-scoring ones to produce the next population, converging to a near-optimal plan.

\section{Results}
We show our proposed EFE-IPP behavior and compare it with other state-of-the-art methods within the synthetic Mars exploration environment. The robot does not have access to the ground truth belief and has to explore the environment, taking measurements/samples to build an accurate information map, limited by the travel length budget.

\begin{figure}[!t]
\includegraphics[width=\textwidth]{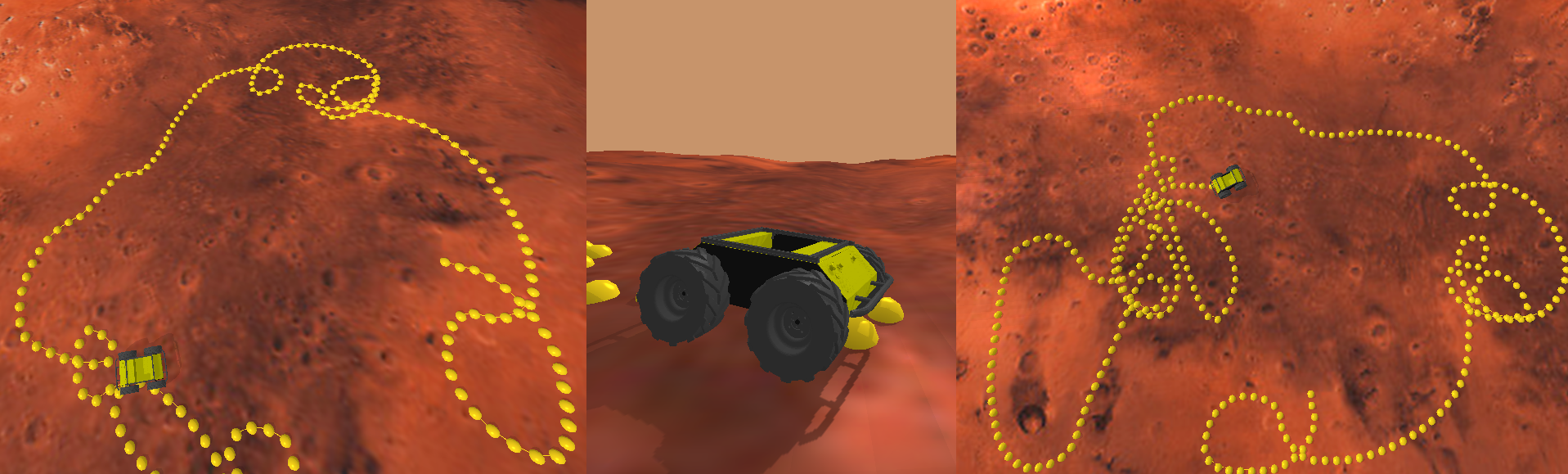}
\caption{Instances of the final (continuous) path executed by the robot navigator in the realistic simulation environment.} \label{fig:robot_path}
\end{figure}

\subsection{Informative Path Planning}

Figure \ref{ipp_result} shows the results of one simulation experiment (details of the simulation setup in Appendix A). The similarities between ground truth and posterior mean of the GP show that our planner was successful in learning the true underlying function. The top right figure shows the decrease in uncertainty around all the locations where measurements were made. The robot initially explores around the map and then concentrates around a region corresponding to the location of the maximum value in the ground truth, indicating heavy exploitation around the end of the path. The bottom panel shows the aEFE objective function, the order of the locations picked for information gathering, and the qualitative plot for the information map. The bottom middle figure shows that the initial points of information gathering are spread throughout the map, while the later points are concentrated around the maximum point. The bottom right figure shows that the map error (RMSE between the GP posterior and the ground truth) and the simple regret (difference between the current maximum found in GP mean and the actual maximum of ground truth) drops as the robot path length increases. The RMSE decreases first while the simple regret decreases later,  indicating an early exploration and later exploitation, showing that our update rule for $\tau^2$ in Equation \ref{eq:tau-schedule} is successful.

\subsection{Planning and Navigation}
Our high level path planner provides the robot navigator (low level controller) with the next path. The controller executes and adapts this path for real-time uncertainties such as changes in terrain height, wrong robot orientation, etc. Figure \ref{fig:robot_path} shows such an instance where the robot adaptively takes continuous paths in the simulation environment. This shows how our framework in Figure \ref{teaser_ipp} can adaptively solve the robot exploration problem in a realistic simulation.

\subsection{Comparative Analysis of the Planner}
\begin{figure}[!t]
\includegraphics[width=\textwidth]{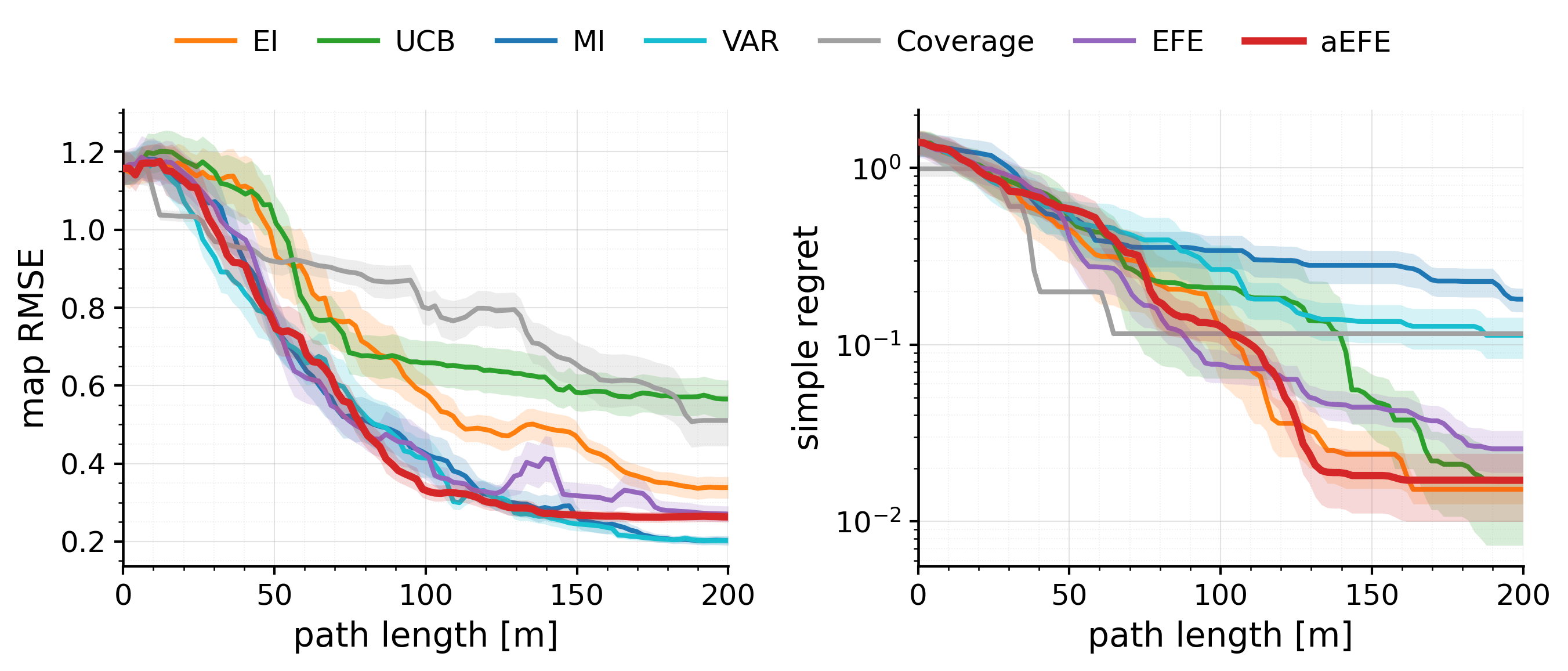}
\caption{The map quality (RMSE) and optimization quality (simple regret) of all methods (results of 20 seeds: mean and SEM) plotted with respect to the robot path length.} \label{fig:compare_Gaussian2D_n20_B200m}
\end{figure}

In this section, we compare our planning algorithm with other methods on a joint optimization and learning problem. Figure \ref{fig:compare_Gaussian2D_n20_B200m} shows the results of 20 seeds of the planner with state of the art algorithms. Expected Improvement and UCB are the most popular acquisition functions from Bayesian Optimization literature, while Mutual Information (MI), VAR (variance $\sigma^2$) and coverage planner \cite{galceran2013survey} are the popular IPP benchmarks. It can be seen that aEFE (our adaptive EFE) is the only method that jointly minimized both map RMSE and simple regret, better than all other methods. EI and UCB are really good at optimization, but they fail to learn the map accurately. MI and VAR are very good at learning the map, but fail to optimize the map. aEFE performs better than EFE towards the end of the budget, showing the effectiveness of our budget-aware $\tau^2$ adaptation.  Coverage planner fails at both optimization and map quality because one measurement per location is clearly suboptimal for information gathering with high measurement noise. However, aEFE strategically refines both optimization and learning through a balanced exploration and exploitation strategy, indicating the competitive nature of our algorithm.

\section{Conclusion}
We presented a budget-aware, adaptive EFE formulation of IPP 
that unifies map learning and optimization under a single, brain-inspired
objective. By scoring continuous trajectories with per-distance EFE and
rolling the belief forward through a fantasy path, the planner
plans non-myopically while respecting a hard movement budget. Across
realizations on a continuous scalar field, our method drives map RMSE and
simple regret down faster than information-theoretic, Bayesian-optimization,
and coverage baselines under identical settings. The same objective recovers a family of qualitatively different search behaviors by changing the preference prior alone, suggesting EFE as a principled, general-purpose objective for robotic information gathering. We see two immediate directions: extending the framework to multi-robot teams that share a belief and budget, and validating the planner on physical platforms where traversability, sensing geometry, and energy budgets enter the same EFE objective.

\begin{credits}
\subsubsection{\ackname} This project was supported by the Dutch Research Council (NWO) under grant AiNed XS Europe (number NGF.1609.243.072). PL was supported by the MetaTool project (Grant agreement 101070940) under the European Innovation Council (EIC) pathfinder program. 

\subsubsection{\discintname}
The authors have no competing interests to declare that are
relevant to the content of this article. 
\end{credits}

\section*{Appendix}
\subsection*{A: Simulation settings}
We evaluate on a $20\,\text{m}\!\times\!20\,\text{m}$ workspace with a hidden
field given by a sum of two Gaussian bumps (peaks at $(5,6)$ and $(5,16)\,\text{m}$),
standardized so that comparisons across objectives are scale-invariant. Each
run begins with $N_0 = 3$ Sobol seed measurements; the robot is then granted a
travel budget of $B = 200\,\text{m}$, with per-segment cap $\ell_{\max} = 5\,\text{m}$
and planning horizon $n = 3$ via-points. Measurements are corrupted by Gaussian
noise of standard deviation $\sigma_n = 0.3$ in the standardized output. The
belief is a \texttt{SingleTaskGP} with an ARD squared-exponential kernel and
output standardization; hyperparameters are refit every $5$ new observations and
updated online otherwise. Inside each plan, via-points are optimized over
$[0, 20]^2$ by differential evolution (\textit{maxiter}~$=12$, \textit{popsize}
multiplier $=5$, no polishing) using the per-length objective with arc floor
$\delta = 0.3\,\text{m}$ to prevent vanishingly short segments.

We compare seven methods on the same fantasy-path framework: EFE (fixed
$\tau^2 = 7$, aspiration $y^{\star} = y_{\text{best}} + 0.1$), aEFE (logistic
schedule $\tau^2: 20 \to 0.6$ over the budget, midpoint $p_0 = 0.4$, steepness
$k = 10$), expected improvement (EI), upper confidence bound (UCB, $\beta = 2.5$),
mutual information (MI), posterior variance (VAR), and a deterministic
lawnmower Coverage baseline ($2\,\text{m}$ margin, $1.5\,\text{m}$ row spacing).
All methods share the same GP, fantasy rollout, optimizer, and budget; only the
per-point objective score differs. We report the mean and standard error of
map RMSE and simple regret over $20$ independent seeds.  

For a qualitative demonstration, we embed our EFE planner in a
\texttt{PyBullet} physics environment that approximates Martian conditions: a procedurally generated $128\!\times\!128$ heightfield terrain (extending a
$3\,\text{m}$ margin beyond the planner's domain so the rover never reaches a cliff), textured with a publicly available albedo map of Mars, and simulated
under Mars surface gravity ($3.71\,\text{m/s}^2$). The robot is a Husky skid-steer rover driven between via-points by a proportional heading-and-speed controller; (synthetic) measurements are taken at the rover's actual
stopped pose, and budget consumption is measured by real traversed distance rather than spline arc length. The GP information map is over a 2D top view projection of the 3D Mars surface.

%
%
%
\bibliographystyle{splncs04}
\bibliography{mybib}

@ARTICLE{meera2026curvature,
  author={Anil Meera, Ajith and Kouw, Wouter M.},
  journal={IEEE Control Systems Letters}, 
  title={Curvature-Aware Expected Free Energy as an Acquisition Function for Bayesian Optimization}, 
  year={2026},
  volume={10},
  number={},
  pages={1693-1698},
  doi={10.1109/LCSYS.2026.3708556}}

@article{friston2010free,
  title={The free-energy principle: a unified brain theory?},
  author={Friston, Karl},
  journal={Nature Reviews Neuroscience},
  volume={11},
  number={2},
  pages={127--138},
  year={2010},
  publisher={Nature publishing group}
}

@article{da2020active,
  title={Active inference on discrete state-spaces: A synthesis},
  author={Da Costa, Lancelot and Parr, Thomas and Sajid, Noor and Veselic, Sebastijan and Neacsu, Victorita and Friston, Karl},
  journal={Journal of Mathematical Psychology},
  volume={99},
  pages={102447},
  year={2020},
  publisher={Elsevier}
}

@book{parr2022active,
  title={Active inference: the free energy principle in mind, brain, and behavior},
  author={Parr, Thomas and Pezzulo, Giovanni and Friston, Karl J},
  year={2022},
  publisher={MIT Press}
}

@article{hollinger2014sampling,
  title={Sampling-based robotic information gathering algorithms},
  author={Hollinger, Geoffrey A and Sukhatme, Gaurav S},
  journal={International Journal of Robotics Research},
  volume={33},
  number={9},
  pages={1271--1287},
  year={2014}
}

@article{chen2024adaptive,
  title={Adaptive robotic information gathering via non-stationary Gaussian processes},
  author={Chen, Weizhe and Khardon, Roni and Liu, Lantao},
  journal={International Journal of Robotics Research},
  volume={43},
  number={4},
  pages={405--436},
  year={2024}
}

@book{williams2006gaussian,
  title={Gaussian processes for machine learning},
  author={Williams, Christopher KI and Rasmussen, Carl Edward},
  year={2006},
  publisher={MIT Press}
}

@article{shahriari2015taking,
  title={Taking the human out of the loop: A review of Bayesian optimization},
  author={Shahriari, Bobak and Swersky, Kevin and Wang, Ziyu and Adams, Ryan P and De Freitas, Nando},
  journal={Proceedings of the IEEE},
  volume={104},
  number={1},
  pages={148--175},
  year={2015},
  publisher={IEEE}
}

@article{kaplan2018planning,
  title={Planning and navigation as active inference},
  author={Kaplan, Raphael and Friston, Karl J},
  journal={Biological Cybernetics},
  volume={112},
  number={4},
  pages={323--343},
  year={2018},
  publisher={Springer}
}

@article{ccatal2021robot,
  title={Robot navigation as hierarchical active inference},
  author={{\c{C}}atal, Ozan and Verbelen, Tim and Van de Maele, Toon and Dhoedt, Bart and Safron, Adam},
  journal={Neural Networks},
  volume={142},
  pages={192--204},
  year={2021},
  publisher={Elsevier}
}

@article{lanillos2021active,
  title={Active inference in robotics and artificial agents: Survey and challenges},
  author={Lanillos, Pablo and Meo, Cristian and Pezzato, Corrado and Meera, Ajith Anil and Baioumy, Mohamed and Ohata, Wataru and Tschantz, Alexander and Millidge, Beren and Wisse, Martijn and Buckley, Christopher L and others},
  journal={arXiv preprint arXiv:2112.01871},
  year={2021}
}

@article{da2022active,
  title={How active inference could help revolutionise robotics},
  author={Da Costa, Lancelot and Lanillos, Pablo and Sajid, Noor and Friston, Karl and Khan, Shujhat},
  journal={Entropy},
  volume={24},
  number={3},
  pages={361},
  year={2022},
  publisher={MDPI}
}

@inproceedings{de2025navigation,
  title={Navigation and Exploration with Active Inference: from Biology to Industry},
  author={de Tinguy, Daria and Verbelen, Tim and Dhoedt, Bart},
  booktitle={International Workshop on Active Inference},
  pages={331--347},
  year={2025},
  organization={Springer}
}

@inproceedings{de2024exploring,
  title={Exploring and learning structure: active inference approach in navigational agents},
  author={de Tinguy, Daria and Verbelen, Tim and Dhoedt, Bart},
  booktitle={International Workshop on Active Inference},
  pages={105--118},
  year={2024},
  organization={Springer}
}

@article{nozari2022active,
  title={Active inference integrated with imitation learning for autonomous driving},
  author={Nozari, Sheida and Krayani, Ali and Marin-Plaza, Pablo and Marcenaro, Lucio and Gomez, David Martin and Regazzoni, Carlo},
  journal={IEEE Access},
  volume={10},
  pages={49738--49756},
  year={2022},
  publisher={IEEE}
}

@article{taniguchi2023active,
  title={Active exploration based on information gain by particle filter for efficient spatial concept formation},
  author={Taniguchi, Akira and Tabuchi, Yoshiki and Ishikawa, Tomochika and El Hafi, Lotfi and Hagiwara, Yoshinobu and Taniguchi, Tadahiro},
  journal={Advanced Robotics},
  volume={37},
  number={13},
  pages={840--870},
  year={2023},
  publisher={Taylor \& Francis}
}

@inproceedings{wakayama2023active,
  title={Active inference for autonomous decision-making with contextual multi-armed bandits},
  author={Wakayama, Shohei and Ahmed, Nisar},
  booktitle={IEEE International Conference on Robotics and Automation},
  pages={7916--7922},
  year={2023}
}

@inproceedings{meera2019obstacle,
  title={Obstacle-aware adaptive informative path planning for uav-based target search},
  author={Meera, Ajith Anil and Popovi{\'c}, Marija and Millane, Alexander and Siegwart, Roland},
  booktitle={IEEE International Conference on Robotics and Automation},
  pages={718--724},
  year={2019}
}

@article{popovic2020informative,
  title={An informative path planning framework for UAV-based terrain monitoring},
  author={Popovi{\'c}, Marija and Vidal-Calleja, Teresa and Hitz, Gregory and Chung, Jen Jen and Sa, Inkyu and Siegwart, Roland and Nieto, Juan},
  journal={Autonomous Robots},
  volume={44},
  number={6},
  pages={889--911},
  year={2020},
  publisher={Springer}
}

@article{friston2015active,
  title={Active inference and epistemic value},
  author={Friston, Karl and Rigoli, Francesco and Ognibene, Dimitri and Mathys, Christoph and Fitzgerald, Thomas and Pezzulo, Giovanni},
  journal={Cognitive neuroscience},
  volume={6},
  number={4},
  pages={187--214},
  year={2015},
  publisher={Taylor \& Francis}
}

@article{das2010differential,
  title={Differential evolution: A survey of the state-of-the-art},
  author={Das, Swagatam and Suganthan, Ponnuthurai Nagaratnam},
  journal={IEEE transactions on evolutionary computation},
  volume={15},
  number={1},
  pages={4--31},
  year={2010},
  publisher={Ieee}
}

@article{galceran2013survey,
  title={A survey on coverage path planning for robotics},
  author={Galceran, Enric and Carreras, Marc},
  journal={Robotics and Autonomous systems},
  volume={61},
  number={12},
  pages={1258--1276},
  year={2013},
  publisher={Elsevier}
}

@article{srinivas2012information,
  title={Information-theoretic regret bounds for gaussian process optimization in the bandit setting},
  author={Srinivas, Niranjan and Krause, Andreas and Kakade, Sham M and Seeger, Matthias W},
  journal={IEEE transactions on information theory},
  volume={58},
  number={5},
  pages={3250--3265},
  year={2012},
  publisher={IEEE}
}

@article{hitz2017adaptive,
  title={Adaptive continuous-space informative path planning for online environmental monitoring},
  author={Hitz, Gregory and Galceran, Enric and Garneau, Marie-{\`E}ve and Pomerleau, Fran{\c{c}}ois and Siegwart, Roland},
  journal={Journal of Field Robotics},
  volume={34},
  number={8},
  pages={1427--1449},
  year={2017},
  publisher={Wiley Online Library}
}

@inproceedings{bai2016information,
  title={Information-theoretic exploration with Bayesian optimization},
  author={Bai, Shi and Wang, Jinkun and Chen, Fanfei and Englot, Brendan},
  booktitle={2016 IEEE/RSJ International Conference on Intelligent Robots and Systems (IROS)},
  pages={1816--1822},
  year={2016},
  organization={IEEE}
}

@article{friston2021sophisticated,
  title={Sophisticated inference},
  author={Friston, Karl and Da Costa, Lancelot and Hafner, Danijar and Hesp, Casper and Parr, Thomas},
  journal={Neural Computation},
  volume={33},
  number={3},
  pages={713--763},
  year={2021},
  publisher={MIT Press One Rogers Street, Cambridge, MA 02142-1209, USA journals-info~…}
}

@article{fountas2020deep,
  title={Deep active inference agents using Monte-Carlo methods},
  author={Fountas, Zafeirios and Sajid, Noor and Mediano, Pedro and Friston, Karl},
  journal={Advances in neural information processing systems},
  volume={33},
  pages={11662--11675},
  year={2020}
}

@article{tschantz2020reinforcement,
  title={Reinforcement learning through active inference},
  author={Tschantz, Alexander and Millidge, Beren and Seth, Anil K and Buckley, Christopher L},
  journal={arXiv preprint arXiv:2002.12636},
  year={2020}
}

\end{document}